\documentclass[letterpaper, 10 pt, conference]{ieeeconf}

\IEEEoverridecommandlockouts

\usepackage{graphics}
\usepackage{epsfig}
\usepackage{amsmath}
\usepackage{amssymb}
\usepackage{float}
\usepackage{xcolor}
\usepackage{pifont}
\usepackage{threeparttable}
\usepackage{mathtools}
\usepackage[ruled,vlined]{algorithm2e}

\usepackage{graphicx}
\usepackage{subcaption}
\usepackage{etoolbox}
\apptocmd{\thebibliography}{\setlength{\itemsep}{0pt}}{}{}

\title{\LARGE \bf
Safety-Critical Bilateral Teleoperation for Omnidirectional Aerial Manipulation Using Force-Sensorless Haptic Feedback
}

\author{Yubin Kim$^{1}$, Jinwoo Lee$^{1}$, Yongjun You$^{1}$, H. Jin Kim$^{2}$, and Jeonghyun Byun$^{3}$%
\thanks{This work was supported by the Agency For Defense Development Grant Funded by the Korean Government(UI257004JD)
}
\thanks{$^{1}$ The authors are with the Department of Aerospace Engineering, Seoul National University,  Seoul, South Korea.
        {\tt\small \{oopuppy, jinwoolee0728, dbys34\}@snu.ac.kr}}
\thanks{$^{2}$ The author is with the Department of Aerospace Engineering, Automation and System Research Institute (ASRI), and Institute of Advanced Aerospace Technology (IAAT), Seoul National University,  Seoul, South Korea. {\tt\small hjinkim@snu.ac.kr}}
\thanks{$^{3}$ The author is with the Automation and System Research Institute (ASRI), Seoul National University,  Seoul, South Korea.
{\tt\small quswjdgus97@snu.ac.kr}}}

\makeatletter

\def\@IEEEtablestring{table}

\long\def\@makecaption#1#2{%
\ifx\@captype\@IEEEtablestring%
  \begin{center}{\footnotesize #1}\\{\footnotesize\scshape #2}\end{center}%
  \@IEEEtablecaptionsepspace%
\else
  \@IEEEfigurecaptionsepspace%
  \setbox\@tempboxa\hbox{\small #1.~~ #2}%
  \ifdim \wd\@tempboxa >\hsize%
    \setbox\@tempboxa\hbox{\small #1.~~ }%
    \parbox[t]{\hsize}{\small \noindent\unhbox\@tempboxa#2}%
  \else%
    \ifcenterfigcaptions
      \hbox to\hsize{\small\hfil\box\@tempboxa\hfil}%
    \else
      \hbox to\hsize{\small\box\@tempboxa\hfil}%
    \fi
  \fi
\fi}

\makeatother

\begin{document}

\maketitle
\thispagestyle{empty}
\pagestyle{empty}

\begin{abstract}
This paper presents a safety-critical bilateral teleoperation framework for omnidirectional aerial manipulators that integrates visual and force-sensorless haptic wrench feedback.
Unlike existing approaches that either rely on onboard force/torque sensors or use model-dependent wrench estimates, which may become unreliable under model uncertainties or induce unintended feedback during free-flight, our method implements a hierarchical safety filter based on control barrier functions to avoid such limitations.
The safety filter, being the key contribution, explicitly accounts for tracking errors arising from physical interaction between the aerial manipulator and its surroundings while enforcing thrust limits, a factor overlooked despite its critical importance for flight safety.
This safety filter adjusts the command from the operator to ensure safe and stable aerial manipulation and avoid motor saturation.
The adjustment made by the filter is mapped to haptic feedback, which is intuitive to the operator and conveys information on physical interaction and impending motor saturation.
By actual experiments with a hexarotor-based omnidirectional aerial manipulator, we demonstrate that the proposed method avoids haptic feedback during free-flight, provides directionally consistent feedback under physical interaction, and can be operated for diverse manipulative tasks.
Moreover, an ablation study further shows that the saturation filter improves interaction stability by explicitly preventing motor saturation and informing the operator of corrective actions.
\end{abstract}

\section{INTRODUCTION}

Aerial manipulators (AMs) are a promising platform for remote operations by combining the maneuverability of aerial vehicles with the versatility of attached robotic manipulators. Therefore, there have been numerous studies to utilize their maneuverability, including non-destructive inspection \cite{watson2021dry}, door opening \cite{lee2020aerial}, plug pulling \cite{byun2021stability}, window cleaning \cite{sun2021switchable}, wall painting \cite{susbielle2025pointillism}, and emergency escape by window breaking \cite{fan2026dynamic}.

Although a number of studies have been conducted on the fully autonomous control of AMs \cite{lee2025autonomous, wen2025full}, bilateral teleoperation remains necessary to leverage human-in-the-loop adaptability, especially in previously unknown environments.
In this regard, there have been several approaches on the bilateral teleoperation of aerial manipulators that leverage sensory feedback such as 2D/3D visual feedback \cite{lee2020visual}, Mixed Reality (MR)/Virtual Reality (VR) interfaces \cite{allenspach2023design, kim2021toward}, or haptic feedback \cite{stramigioli2010novel, gioioso2015force}. These works have paved the way towards precise, safe, and transparent teleoperation of AMs with adequate sensory feedback.
\begin{figure}[!t] 
    \centering  \includegraphics[width=0.9\linewidth]{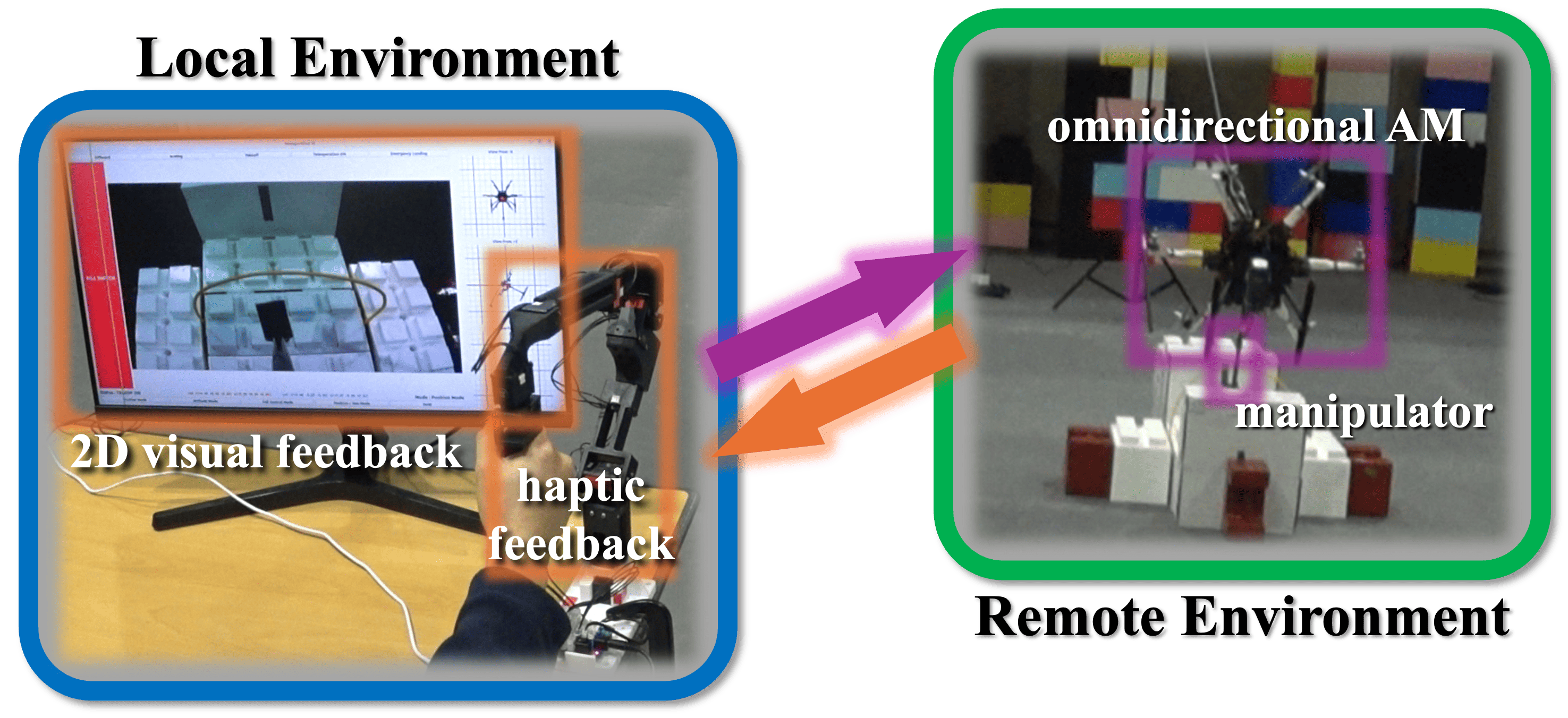}
    \caption{Illustration of our haptic-based bilateral teleoperation for an omnidirectional aerial manipulator (remote environment) teleoperated by a human operator through a haptic master device (local environment).}
    \label{fig: thumbnail}
\end{figure}

In particular, for tasks involving physical interaction, haptic feedback plays an important role in enhancing the transparency of the teleoperation. The straightforward solution is the attachment of the force/torque (FT) sensors on both AMs and haptic devices; however, we aim to avoid this solution to maintain the minimum sensor configuration of our framework. Meanwhile, existing FT-sensorless haptic feedback approaches have typically relied on model-dependent, momentum-based wrench estimation, which might become unreliable under model uncertainties. Furthermore, in terms of transparency, providing the operator with clear and reliable cues of physical interaction is more imperative than accurately measuring or estimating the interaction wrench.
\par
On the other hand, complying with thrust limits is crucial for the safe operation of aerial manipulators but is frequently overlooked.
Moreover, informing the operator of impending motor saturation facilitates safety in performing aerial manipulative tasks.
\par
For that end, we introduce an FT-sensorless bilateral teleoperation framework, requiring no FT sensors on either the AM or the master device, for an omnidirectional aerial manipulator. It employs a hierarchical safety filter that adjusts the commanded pose and generates haptic feedback based on (i) tracking errors induced by physical interaction and (ii) thrust limits, promoting safer and more stable interaction. The feedback remains directionally coherent regardless of the vehicle's attitude, letting the operator intuitively exploit the system's omnidirectionality.

\subsection{Related Works}
In \cite{mohammadi2016cooperative}, a bilateral teleoperation method for an AM using haptic feedback is proposed and validated in simulation, and in \cite{allenspach2022towards}, a seven-degree-of-freedom (DoF) robotic manipulator is used to command the full pose of an AM. A semi-autonomous teleoperation framework for a push-and-slide task using vision and force feedback from the vehicle is introduced in \cite{d2025semi}. However, these works rely on FT sensors to directly measure and render interaction wrenches to the operator.

To maintain the minimal sensor configuration of the teleoperation frameworks, there exist several studies on bilateral teleoperation of AMs using haptic feedback that do not require FT sensors on the aerial vehicle's end-effector. In \cite{kim2020human}, aerial drilling via teleoperation is presented, where interaction forces are inferred using current measurements from a servo-actuated drill manipulator mounted on the aerial vehicle. However, this approach is limited to estimating only the vertical force acting on the drill bit. In \cite{byun2024haptic} and \cite{mellet2025evaluation}, interaction wrenches acting on AMs are estimated using momentum-based wrench estimators. Nevertheless, due to model uncertainties such as mass, inertia, and thrust-model errors, the estimated wrench may drift or fluctuate even when there is no physical interaction. Meanwhile, \cite{coelho2020whole} and \cite{coelho2021whole} utilize the control wrench computed by the tracking controller of the AM as haptic feedback, rather than directly estimating the interaction wrench.
However, this method may lead to unintended residual haptic feedback even when the operator has no intention of interaction, since a nonzero tracking error may persist unless the operator intentionally commands to reduce it.
Moreover, motor saturation has rarely been incorporated into bilateral teleoperation of AMs, despite being critical for flight safety during physical interaction.

Finally, fully omnidirectional aerial manipulation has also been explored in teleoperation. For example, \cite{li2025six} and \cite{kaneko2026teleop} propose a teleoperation method for fully omnidirectional aerial manipulation and validates it by turning a valve while maintaining vertical flight. However, the hand-worn interface used in this work focuses on motion-based teleoperation and does not provide haptic wrench feedback, which may limit the achievable transparency during physical interaction. 
\par
To the best of our knowledge, no previous work simultaneously incorporates: (i) motor thrust limits, (ii) FT-sensorless haptic feedback, and (iii) fully omnidirectional aerial manipulation.

\subsection{Contributions}
To summarize, the main contributions of this paper are as follows:
\begin{itemize}
    \item The first bilateral teleoperation framework for omnidirectional aerial manipulation using vision and haptic feedback that explicitly accounts for motor thrust limits.
    \item The design of a safety filter that adjusts the desired pose of the AM based on the tracking error and motor thrust limits; this filter guarantees thrust feasibility while preserving task performance as much as possible, thereby preventing motor saturation during physical interaction.
    \item The design of an FT-sensorless haptic feedback method reflecting the adjustments to the commanded pose; this method eliminates the need of FT sensors while preserving informative interaction cues.
    \item Hardware demonstrations of the proposed framework through actual experiments, including both static and dynamic physical interaction.
\end{itemize}

\subsection{Outline}
Section II outlines the overall structure of the bilateral teleoperation framework.
Section III describes the control method implemented in the master device, while Section IV details the safety filter used for generating haptic feedback from the aerial manipulator.
In Section V, experimental results validating the performance of the proposed framework are presented.

\begin{figure}[!t]
    \centering
\includegraphics[width=0.9\linewidth]{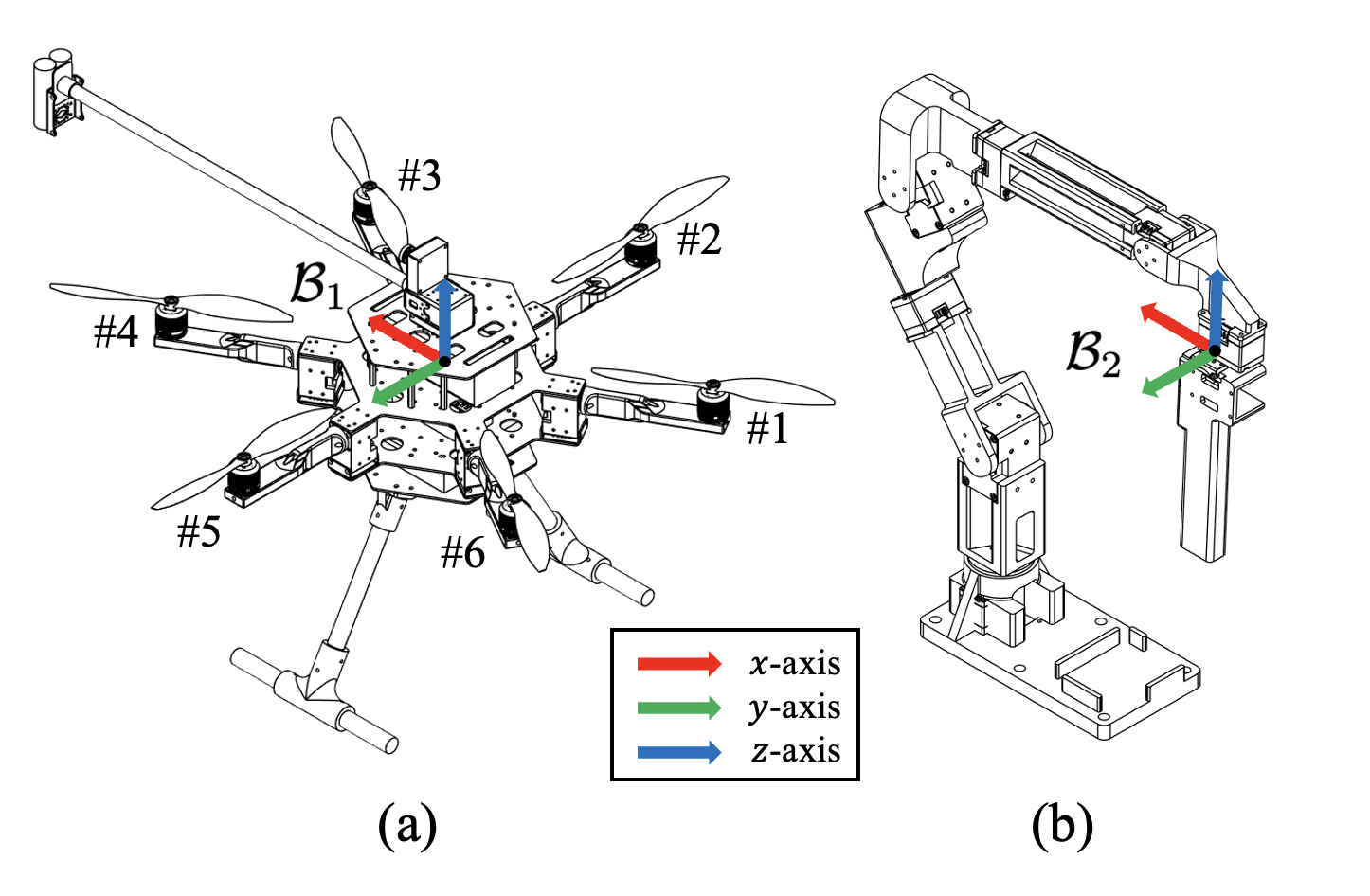}
\vspace{-2mm}
    \caption{Three-dimensional models of (a) the AM and (b) the master device, including their corresponding body frames, with motor indices labeled on the AM.}
    \label{fig:modeling}
\end{figure}

\section{Bilateral Teleoperation Framework}
\label{sec2}
\subsection{Notations and Coordinate Frames}
\par
The symbol $\boldsymbol{e_3} \in \mathbb{R}^3$ is used to represent the unit vector $[0; 0; 1]$, while $\boldsymbol{I}_n$ indicates the $n \times n$ identity matrix.
Estimates of physical quantities are distinguished by the hat symbol $\hat{(\cdot)}$.
Also, the rotation from ${\mathcal{F}_1}$ to ${\mathcal{F}_2}$ is denoted by rotation matrix $\prescript{\mathcal{F}_2}{\mathcal{F}_1}{\boldsymbol{R}} \in SO(3)$.

\par
The 2-norm of a vector is represented by $\lVert \cdot \rVert$, while the weighted 2-norm with weighting matrix $\boldsymbol{W}$ is denoted by $\lVert \cdot \rVert_{\boldsymbol{W}}$.
The Moore-Penrose pseudoinverse of a matrix $\boldsymbol{X}$ is written as $\boldsymbol{X}^{\dagger}$.
For vectors $\boldsymbol{a}, \boldsymbol{b} \in \mathbb{R}^3$, the skew-symmetric operator $(\cdot)_{\times}$ is defined such that $\boldsymbol{a}_{\times} \boldsymbol{b} = \boldsymbol{a} \times \boldsymbol{b}$.
The inverse of the skew-symmetric operation is represented by the vee operator $(\cdot)^\vee$.

\par
Relevant coordinate frames include the world inertial frame ${\mathcal{W}}$, the AM body frame ${\mathcal{B}_1}$, and the master device end-effector frame ${\mathcal{B}_2}$, which are depicted in Fig.~\ref{fig:modeling}.
\subsection{Outline}

\begin{figure*}[!t]
    \centering
    \includegraphics[width=1.0\textwidth]{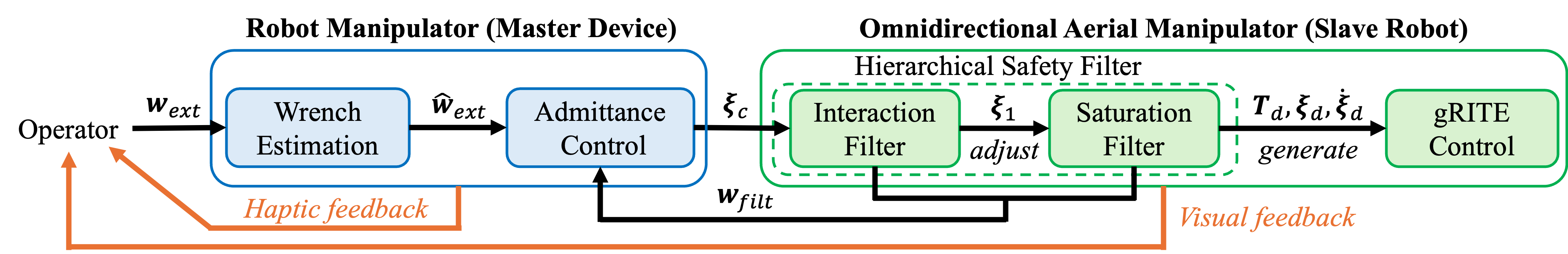}
    \caption{Outline of the teleoperation framework.}
    \vspace{-4mm}
    \label{fig: teleoperation framework}
\end{figure*}

\par
The overall architecture of the proposed bilateral teleoperation framework is illustrated in Fig. \ref{fig: teleoperation framework}.
The framework is composed of two primary components: a robotic manipulator that functions as the master device of the teleoperation system and an omnidirectional aerial manipulator with tiltable rotor arms that serves as the slave robot.

\par
The master device is driven by an admittance controller with a combined wrench input consisting of the estimated external wrench at the end-effector, $\hat{\boldsymbol{w}}_{ext} = [\hat{\boldsymbol{f}}_{ext}; \hat{\boldsymbol{\tau}}_{ext}] \in \mathbb{R}^6$, and the safety-filter feedback wrench, $\boldsymbol{w}_{filt} = [\boldsymbol{f}_{filt}; \boldsymbol{\tau}_{filt}] \in \mathbb{R}^6$. Here, $\hat{\boldsymbol{w}}_{ext}$ denotes an estimate of the true external wrench $\boldsymbol{w}_{ext} = [\boldsymbol{f}_{ext}; \boldsymbol{\tau}_{ext}] \in \mathbb{R}^6$ applied to the end-effector. The resulting deviation of the current end-effector pose from its neutral pose is then mapped to the twist command $\boldsymbol{\xi}_c\in \mathbb{R}^6$ for the AM.

\par
On the AM side, the twist command translated from the master device is processed through a hierarchical safety filter.
In the first stage (the interaction filter), the command twist is modified to satisfy tracking error constraints, in effect securing command compliance with the AM's interactions and generating a feedback wrench that reflects this interaction with the environment.
The second stage (the saturation filter) further adjusts the command to ensure that motor thrust limits are not violated, producing a corresponding feedback wrench associated with motor saturation.
The wrenches from both stages are combined to form the overall feedback wrench, $\boldsymbol{w}_{filt}$, which is translated back to the master device.

The command modified by the hierarchical safety filter is integrated with the current pose of the AM to generate the desired pose $\boldsymbol{T}_d \in SE(3)$, desired twist $\boldsymbol{\xi}_d \in \mathbb{R}^6$, and desired spatial acceleration $\dot{\boldsymbol{\xi}}_d \in \mathbb{R}^6$.
Based on $\boldsymbol{T}_d$, $\boldsymbol{\xi}_d$, and $\dot{\boldsymbol{\xi}}_d$, the gRITE (Geometric Robust Integration Integral of the tanh of the Error) controller \cite{lee2025autonomous} together with geometric allocation \cite{kamel2018voliro} computes the thrust magnitudes and rotor tilt angles necessary to track the desired state.

\section{Master Device}
\label{sec3}
\subsection{Dynamics}
\par
The robotic manipulator that acts as the master device of the teleoperation framework can be modeled as a single open chain of rigid bodies connected by $n_m$ revolute joints.
Let $\boldsymbol{q} \in \mathbb{R}^{n_m}$ denote the joint angles; then the Euler-Lagrange equation of motion of the master device is derived as follows:
\begin{equation}
\label{eqn: robotic arm dynamics}
\boldsymbol{M}(\boldsymbol{q}) \ddot{\boldsymbol{q}} + \boldsymbol{C}(\boldsymbol{q}, \dot{\boldsymbol{q}}) \dot{\boldsymbol{q}} + \boldsymbol{g}(\boldsymbol{q}) = \boldsymbol{\tau}^{[jnt]} + J_{ee}^T(\boldsymbol{q}) \boldsymbol{w}_{ext}
\end{equation}
where $\boldsymbol{M}(\boldsymbol{q})$ and $\boldsymbol{C}(\boldsymbol{q}, \dot{\boldsymbol{q}}) \in \mathbb{R}^{n_m \times n_m}$ are the inertial matrix and the matrix containing Coriolis and centrifugal effects factorized by Christoffel symbols, respectively, $\boldsymbol{g}(\boldsymbol{q}) \in \mathbb{R}^{n_m}$ is the generalized gravitational vector, and $\boldsymbol{\tau}^{[jnt]} \in \mathbb{R}^{n_m}$ are the command joint torques.
The geometric Jacobian of the end-effector, $\boldsymbol{J}_{ee}(\boldsymbol{q}) \in \mathbb{R}^{6 \times n_m}$, relates the end-effector twist $\boldsymbol{\xi}^{[ee]} \in \mathbb{R}^6$ to the joint velocities $\dot{\boldsymbol{q}} \in \mathbb{R}^{n_m}$ as $\boldsymbol{\xi}^{[ee]} = \boldsymbol{J}_{ee}(\boldsymbol{q}) \dot{\boldsymbol{q}}$.

\subsection{External Wrench Estimation}
\par
For the admittance control of the master device's end-effector, the external wrench $\boldsymbol{w}_{ext}$ exerted by the operator is required.
To avoid incorporating FT sensors into the framework, the external wrench is estimated following the approach in \cite{deluca2006collision}, as described below:
\begin{multline} \label{eqn: equation for r}
\hat{\boldsymbol{w}}_{ext} = (\boldsymbol{J}_{ee}^T(\boldsymbol{q}))^\dagger \Big[ \boldsymbol{K}_E \big\{ \boldsymbol{p}(t) - \boldsymbol{p}(0) \\ -\int_0^t (\boldsymbol{\tau}^{[jnt]} + \boldsymbol{C}^T(\boldsymbol{q}, \dot{\boldsymbol{q}}) - \boldsymbol{g}(\boldsymbol{q}) + \boldsymbol{J}_{ee}^T(\boldsymbol{q}) \hat{\boldsymbol{w}}_{ext}) ds \big\} \Big]
\end{multline}
where $\boldsymbol{K}_E \in \mathbb{R}^{n_m \times n_m}$ is a positive diagonal gain matrix and $\boldsymbol{p} = \boldsymbol{M}(\boldsymbol{q}) \dot{\boldsymbol{q}}$ is the generalized momentum.

\subsection{Admittance Control}
\par
As introduced in \cite{allenspach2022towards}, we utilize an admittance controller combined with a low-level joint controller to guarantee compliance with respect to the human operator's interaction wrench and haptic feedback from the AM.
To this end, given the estimated external wrench $\hat{\boldsymbol{w}}_{ext}$ and the feedback wrench $\boldsymbol{w}_{filt}$, we first formulate the dynamics of the desired end-effector pose as follows:
\begin{equation}
\begin{aligned} \label{eqn: admittance control}
&\boldsymbol{M}_t \ddot{\boldsymbol{p}}^{[ee]}_d + \boldsymbol{C}_t \dot{\boldsymbol{p}}^{[ee]}_d + \boldsymbol{P}_t (\boldsymbol{p}^{[ee]}_d - \boldsymbol{p}^{[ee]}_0) = \hat{\boldsymbol{f}}_{ext} + \boldsymbol{f}_{filt} \\ 
&\boldsymbol{M}_r \dot{\boldsymbol{\Omega}}^{[ee]}_d + \boldsymbol{C}_r \boldsymbol{\Omega}^{[ee]}_d + \boldsymbol{P}_r \boldsymbol{e}^{[ee]}_R = \prescript{\mathcal{B}_2}{\mathcal{W}}{\boldsymbol{R}}^T (\hat{\boldsymbol{\tau}}_{ext} + \boldsymbol{\tau}_{filt})
\end{aligned}
\end{equation}
where
\begin{equation*}
\boldsymbol{e}_R^{[ee]} = \tfrac{1}{2} (\prescript{\mathcal{B}_2}{\mathcal{W}}{\boldsymbol{R}}_0^T \prescript{\mathcal{B}_2}{\mathcal{W}}{\boldsymbol{R}}_d - \prescript{\mathcal{B}_2}{\mathcal{W}}{\boldsymbol{R}}_d^T \prescript{\mathcal{B}_2}{\mathcal{W}}{\boldsymbol{R}}_0)
\end{equation*}
with $\prescript{\mathcal{B}_2}{\mathcal{W}}{\boldsymbol{R}}_d$ and $\prescript{\mathcal{B}_2}{\mathcal{W}}{\boldsymbol{R}}_0\in SO(3)$ denoting the rotation matrices corresponding to the end-effector's desired and neutral orientation.
Additionally, $\boldsymbol{p}^{[ee]}_d$ and $\boldsymbol{\Omega}^{[ee]}_d \in \mathbb{R}^3$ represent the desired end-effector position and desired angular velocity, respectively, and $\boldsymbol{p}_0 \in \mathbb{R}^3$ is the neutral position of the end-effector. The design matrices $\boldsymbol{M}_t, \boldsymbol{C}_t, \boldsymbol{P}_t, \boldsymbol{M}_r, \boldsymbol{C}_r, \boldsymbol{P}_r \in \mathbb{R}^{3 \times 3}$ are selected so that the end-effector exhibits the behavior of a prescribed mass-spring-damper system anchored at its neutral pose.

\par
The joint angles corresponding to the desired end-effector pose obtained from (\ref{eqn: admittance control}) are computed by an iterative inverse kinematics (IK) algorithm with null-space projection
to manage the kinematic redundancy of the master device.

\subsection{Master Command Generation}
\par
The master twist command, $\boldsymbol{\xi}_c$, is computed by mapping the deviation of the master device's end-effector pose with respect to its neutral configuration.
In our framework, we used a linear deadband mapping shown below:
\begin{multline} \label{eqn: generation of master commands}
\xi_{c, i} = 
\begin{cases}
0 & \lvert e^{[ee]}_i \rvert \le l_i \\
K_{M,i} (e^{[ee]}_i - l_i \operatorname{sgn}(e^{[ee]}_i)) & \lvert e^{[ee]}_i \rvert > l_i
\end{cases}
, \\
i \in \{ 1, 2, \cdots, 6 \}
\end{multline}
where $\boldsymbol{e}^{[ee]} = [\boldsymbol{p}_d^{[ee]}-\boldsymbol{p}_0^{[ee]}; \boldsymbol{e}_R^{[ee]}]$, $\boldsymbol{K}_{M} \in \mathbb{R}^{6 \times 6}$ is a positive diagonal gain matrix, and $\boldsymbol{l} \in \mathbb{R}^6$ is a vector containing deadbands for each wrench component.
By introducing deadbands in (\ref{eqn: generation of master commands}), we ensure that a nonzero twist command is generated only when the master device's end-effector pose deviates sufficiently from the neutral configuration. 
Furthermore, $\boldsymbol{\xi}_c$ is selectively gated according to the active control mode. Specifically, only the relevant translational and/or rotational components are transmitted to the slave in the position-only, attitude-only, full 6-DoF, and position+yaw modes. Hereafter, $\boldsymbol{\xi}_c$ refers to this gated command.

\section{Omnidirectional Aerial Manipulator}
\label{sec4}
\par
In this section, we introduce our two-stage hierarchical safety filter that adjusts the command twist of the AM, $\boldsymbol{\xi}_c$, to (i) provide interaction that is consistent with constraints imposed by the AM's surroundings and (ii) prevent motor saturation.
The first stage is an interaction filter that adjusts the command twist sent from the master device such that tracking-error constraints are satisfied, thus providing a command consistent with interactions made with the surroundings.
The saturation filter of the second stage further modifies the command adjusted by the interaction filter so that thrust limits are met.

\par
The interaction filter generates a haptic feedback wrench reflecting the aerial manipulator's interaction with its surroundings, whereas the saturation filter produces a feedback wrench associated with motor saturation.
Both filters are formulated as quadratic programs (QPs) subject to control barrier function (CBF) constraints, as detailed below.

\subsection{Preliminary: Affine Mappings in Aerial Manipulator Dynamics}
\par
Prior to formulating the interaction and saturation filters, the closed-loop error dynamics of our omnidirectional aerial manipulator equipped with tiltable rotors need to be derived.

\par
According to \cite{lee2025autonomous}, the control wrench $\boldsymbol{w}$ is locally affine to the desired spatial acceleration $\dot{\boldsymbol{\xi}}_d$:
\begin{equation} \label{eqn: affine mapping 1}
\boldsymbol{w} = \boldsymbol{w}_0 + \boldsymbol{G} \dot{\boldsymbol{\xi}}_d
\end{equation}
where $\boldsymbol{w}_0$ collects all terms independent of $\dot{\boldsymbol{\xi}}_d$ and $\boldsymbol{G} \in \mathbb{R}^{6 \times 6}$ is known locally.
Actuator allocation \cite{kamel2018voliro} produces planar thrusts for each rotor from the control wrench:
\begin{equation} \label{eqn: allocation}
[ \boldsymbol{T}_1; \  \cdots; \ \boldsymbol{T}_{n_r}]
= \boldsymbol{B}^\dagger \boldsymbol{w}, \qquad
\boldsymbol{T}_i \triangleq
\begin{bmatrix}
T_i \cos \theta_i \\ T_i \sin \theta_i
\end{bmatrix}
\in \mathbb{R}^2
\end{equation}
where $n_r$ is the number of rotors; $T_i, \theta_i \in \mathbb{R}$ are rotor thrust magnitudes and angles, respectively; and $\boldsymbol{B} \in \mathbb{R}^{6 \times 2n_r}$ is the thrust allocation matrix.
By substituting (\ref{eqn: allocation}) for (\ref{eqn: affine mapping 1}), the thrust vector of rotor $i$ becomes affine in $\dot{\boldsymbol{\xi}}_d$:
\begin{equation}
\begin{aligned} \label{eqn: affine mapping 2}
\boldsymbol{T}_i &= \boldsymbol{A}_i \dot{\boldsymbol{\xi}}_d + \boldsymbol{b}_i, \qquad \\
\boldsymbol{A}_i &\triangleq \boldsymbol{E}_i \boldsymbol{B}^\dagger \boldsymbol{G} \in \mathbb{R}^{2 \times 6}, \qquad 
\boldsymbol{b}_i \triangleq \boldsymbol{E}_i \boldsymbol{B}^\dagger \boldsymbol{w}_0 \in \mathbb{R}^2
\end{aligned}
\end{equation}
Here, $\boldsymbol{E}_i \in \mathbb{R}^{2 \times 2n_r}$ selects the two rows associated with the $i$th rotor.

\subsection{1st Stage: The Interaction Filter}
\par
To enforce compliance with constraints imposed by the AM's interaction with its surroundings, the interaction filter modifies the master command twist $\boldsymbol{\xi}_c$ into twist $\boldsymbol{\xi}_1 \triangleq [\boldsymbol{v}_1; \boldsymbol{\Omega}_1] \in \mathbb{R}^6$, where $\boldsymbol{v}_1$ and $\boldsymbol{\Omega}_1$ are decision variables.

\subsubsection{Error dynamics}
\par
Suppose that $\boldsymbol{p}\in \mathbb{R}^3$ and $\prescript{\mathcal{B}_1}{\mathcal{W}}{\boldsymbol{R}} \in SO(3)$ represent the current pose of the aerial manipulator, while $\boldsymbol{p}_d \in \mathbb{R}^3$ and $\prescript{\mathcal{B}_1}{\mathcal{W}}{\boldsymbol{R}_d} \in SO(3)$ represent the desired pose computed by integrating master command twist $\boldsymbol{\xi}_c$ to the current pose.
The associated tracking errors are defined as
\begin{equation} \label{eqn: errors for error filter}
\boldsymbol{e}_p \triangleq \boldsymbol{p} - \boldsymbol{p}_d, \quad 
\boldsymbol{e}_R \triangleq \tfrac{1}{2} (\boldsymbol{\Psi}^T - \boldsymbol{\Psi})^\vee, \quad  
\boldsymbol{e}_\Omega \triangleq \boldsymbol{\Omega} - \boldsymbol{\Psi}\boldsymbol{\Omega}_1
\end{equation}
where $\boldsymbol{\Psi} \triangleq \prescript{\mathcal{B}_1}{\mathcal{W}}{\boldsymbol{R}}^T \prescript{\mathcal{B}_1}{\mathcal{W}}{\boldsymbol{R}}_d$.

\par
The position error $\boldsymbol{e}_p$ evolves as
\begin{equation} \label{eqn: error dynamics 1}
\dot{\boldsymbol{e}}_p = \boldsymbol{v} - \boldsymbol{v}_1,
\end{equation}
where $\boldsymbol{v} \in \mathbb{R}^3$ is the current linear velocity of the aerial manipulator.
Furthermore, the time derivative of $\boldsymbol{e}_R$ is related to the angular-velocity error $\boldsymbol{e}_\Omega$ defined in (\ref{eqn: errors for error filter}) by the kinematic relationship
\begin{equation} \label{eqn: error dynamics 2}
\dot{\boldsymbol{e}}_R = \boldsymbol{\mathcal{E}}(\boldsymbol{\Psi}) \boldsymbol{e}_\Omega, \quad
\boldsymbol{\mathcal{E}}(\boldsymbol{\Psi}) \triangleq \tfrac{1}{2}(\operatorname{tr}(\boldsymbol{\Psi}) \boldsymbol{I}_3 - \boldsymbol{\Psi}^T)
\end{equation}
It is noted that the decision variables, $\boldsymbol{v}_1$ and $\boldsymbol{\Omega}_1$, enter our formulation through the error dynamics (\ref{eqn: error dynamics 1}) and (\ref{eqn: error dynamics 2}).

\subsubsection{Filter Design}
\par
Bounds on the position and attitude tracking errors are enforced via soft CBFs.
For position errors, we let the admissible position-error bounds be
\begin{equation} \label{eqn: position error bounds}
e_{p, i} \in [-e_{p,i}^{[M]}, e_{p, i}^{[M]}], \qquad
i \in \{x, y, z\}.
\end{equation}
Therefore, the control barrier functions with respect to the position errors are defined as follows:
\begin{equation} \label{eqn: position error CBF def}
h_{p, i}^-(\boldsymbol{e}_p) \triangleq e_{p, i} + e_{p, i}^{[M]}, \qquad
h_{p, i}^+(\boldsymbol{e}_p) \triangleq e_{p, i}^{[M]} - e_{p, i}.
\end{equation}
Now consider inequalities
\begin{equation} \label{eqn: bounds on position error CBF}
\dot{h}_{p, i}^{\pm} + \alpha_p h_{p, i}^{\pm} \ge -\delta_{p, i}^{\pm}, \qquad \delta_{p, i}^{\pm} \ge 0 
\end{equation}
where $\alpha_p \in \mathbb{R}$ is a tuning parameter that dictates the convergence rate of the constraint and $\delta_{p, i}^{\pm} \in \mathbb{R}_+$ is a slack variable.
Since $\dot{\boldsymbol{e}}_p = \boldsymbol{v} - \boldsymbol{v}_1$, (\ref{eqn: bounds on position error CBF}) is rearranged with respect to $\boldsymbol{v}_1$ as follows:
\begin{equation}
\begin{aligned} \label{eqn: affine bounds on v_1}
v_{1, i} & \le v_i + \alpha_p (e_{p, i} + e_{p, i}^{[M]}) + \delta_{p, i}^- \\
v_{1, i} & \ge v_i + \alpha_p (e_{p, i} - e_{p, i}^{[M]}) - \delta_{p, i}^+.
\end{aligned}
\end{equation}

\par
Regarding attitude errors, let $\theta_R > 0$ denote the admissible attitude-error radius; then the associated CBF is defined as
\begin{equation} \label{eqn: attitude error CBF def}
h_R(\boldsymbol{e}_R) \triangleq \theta_R^2 - \lVert \boldsymbol{e}_R \rVert^2.
\end{equation}
Using (\ref{eqn: error dynamics 2}), we have
\begin{equation} \label{eqn: dot for h_R}
\dot{h}_R = -2 \boldsymbol{e}_R^T \dot{\boldsymbol{e}}_R = -2 \boldsymbol{e}_R^T \boldsymbol{\mathcal{E}}(\boldsymbol{\Psi}) (\boldsymbol{\Omega} - \boldsymbol{\Psi} \boldsymbol{\Omega}_1).
\end{equation}
Hence, enforcing the condition
\begin{equation} \label{eqn: bounds on attitude error CBF}
\dot{h}_R + \alpha_R h_R \ge -\delta_R, \qquad
\delta_R \ge 0
\end{equation}
where $\alpha_R, \delta_R \in \mathbb{R}$ are a tuning parameter and a slack variable, respectively, an affine inequality of $\boldsymbol{\Omega}_1$ is obtained as follows:
\begin{equation} \label{eqn: affine bounds on Omega_1}
2 \boldsymbol{e}_R^T \boldsymbol{\mathcal{E}}(\boldsymbol{\Psi}) \boldsymbol{\Psi} \boldsymbol{\Omega}_1 \ge 2 \boldsymbol{e}_R^T \boldsymbol{\mathcal{E}}(\boldsymbol{\Psi}) \boldsymbol{\Omega} - \alpha_R h_R - \delta_R.
\end{equation}

\par
Finally, with the CBF inequalities (\ref{eqn: affine bounds on v_1}) and (\ref{eqn: affine bounds on Omega_1}) the interaction filter is constructed as a CBF-QP formulation as follows:
\begin{equation}
\begin{aligned} \label{eqn: QP for error filter}
(\boldsymbol{\xi}_1^*, \boldsymbol{\delta}^*) = \arg \min_{\substack{\boldsymbol{\xi}_1 \\ \boldsymbol{\delta} \ge \boldsymbol{0}}} &\tfrac{1}{2} \lVert \boldsymbol{\xi}_1 - \boldsymbol{\xi}_c \rVert^2_{\boldsymbol{W}_e} + \tfrac{1}{2} \lVert \boldsymbol{\delta} \rVert^2_{\boldsymbol{W}_\delta} \\ 
\text{s.t.} \quad & (\ref{eqn: affine bounds on v_1}), (\ref{eqn: affine bounds on Omega_1})
\end{aligned}
\end{equation}
where $\boldsymbol{\delta} \triangleq [\delta_{p, x}^-, \delta_{p, x}^+, \cdots, \delta_{p, z}^-, \delta_{p, z}^+, \delta_R]^T \in \mathbb{R}^7$, while $\boldsymbol{W}_e \in \mathbb{R}^{6 \times 6}$ and $\boldsymbol{W}_\delta \in \mathbb{R}^{7 \times 7}$ are weights for command deviation and penalization on constraint relaxation, respectively. Here, $\boldsymbol{W}_\delta$ is set small to keep the error bounds loose for interaction feedback.

\subsection{2nd Stage: The Saturation Filter}
\par
The second stage of our hierarchical safety filter, the saturation filter, prevents motor saturation by adjusting the output of the interaction filter by generating a jerk input $\ddot{\boldsymbol{\xi}}_2 \in \mathbb{R}^6$ that complies with motor thrust limits.

\subsubsection{Nominal acceleration/jerk to reach the first-stage velocity reference}
The output twist of the interaction filter, $\boldsymbol{\xi}_1^*$, is used as a reference in the formulation of the saturation filter.
Specifically, the nominal desired acceleration $\dot{{\boldsymbol{\xi}}}_{nom}$ required for reaching the velocity reference is defined as:
\begin{equation} \label{eqn: nominal acceleration}
\dot{{\boldsymbol{\xi}}}_{nom} \triangleq \tfrac{1}{T_{c_1}} (\boldsymbol{\xi}_1^* - \boldsymbol{\xi}_e)
\end{equation}
where $\boldsymbol{\xi}_e$ is the current desired twist and
$T_{c1} > 0$ represents a time constant.
Using (\ref{eqn: nominal acceleration}), the nominal jerk $\ddot{\boldsymbol{\xi}}_{nom}$ is defined as follows:
\begin{equation} \label{eqn: nominal jerk}
\ddot{\boldsymbol{\xi}}_{nom} \triangleq \tfrac{1}{T_{c_2}} (\dot{{\boldsymbol{\xi}}}_{nom} - \dot{\boldsymbol{\xi}}_e)
\end{equation}
where $\dot{\boldsymbol{\xi}}_e$ is the time derivative of $\boldsymbol{\xi}_e$ and $T_{c_2} > 0$ is another time constant.
The saturation filter calculates the jerk $\ddot{\boldsymbol{\xi}}_2$ that minimally deviates from $\ddot{\boldsymbol{\xi}}_{nom}$ while enforcing thrust constraints. 

\subsubsection{Filter Design}
\par
We enforce the hard bound $\lVert \boldsymbol{T}_i \rVert \le T_{i, max}$ for $\forall i \in \{1, \cdots, n_r\}$ by defining the control barrier function concerning thrust limits as:
\begin{equation} \label{eqn: thrust CBF def}
h_{T, i}(\dot{\boldsymbol{\xi}}_2) \triangleq T_{i, max}^2 - \lVert \boldsymbol{T}_i(\dot{\boldsymbol{\xi}}_2) \rVert ^2 .
\end{equation}
From (\ref{eqn: affine mapping 2}), there exists a matrix $\boldsymbol{A}_i \in \mathbb{R}^{2 \times 6}$ such that
\begin{equation}
\dot{\boldsymbol{T}_i} = \boldsymbol{A}_i \ddot{\boldsymbol{\xi}}_2
\end{equation}
and consequently, $\dot{h}_{T, i}$ is derived as follows:
\begin{equation} \label{eqn: h_dot for thrust}
\dot{h}_{T, i} = -2 \boldsymbol{T}_i^T \dot{\boldsymbol{T}}_i = -2(\boldsymbol{A}_i\dot{\boldsymbol{\xi}} + \boldsymbol{b}_i)^T \boldsymbol{A}_i \ddot{\boldsymbol{\xi}}_2 .
\end{equation}
Therefore, with a class-$\mathcal{K}$ function $\alpha_{T, i} h_{T, i} - \tfrac{\gamma_{T, i}}{h_{T, i}} \ge 0 \ \forall t \ge 0$, the CBF condition is formulated as follows:
\begin{equation} \label{eqn: linear ineq for thrust h}
\dot{h}_{T, i} + \alpha_{T, i} h_{T, i} - \tfrac{\gamma_{T,i}}{h_{T,i}} \ge 0,
\end{equation}
with tuning parameters $\alpha_{T, i}$ and $\gamma_{T,i} \in \mathbb{R}$. Then, (\ref{eqn: linear ineq for thrust h}) yields a linear inequality with respect to decision variable $\ddot{\boldsymbol{\xi}}_2$ as:
\begin{equation} \label{eqn: linear ineq for QP}
\begin{split}
2 \boldsymbol{T}_i^T \boldsymbol{A_i} \ddot{\boldsymbol{\xi}}_2 \le  \\ 
    \alpha_{T, i} &(T_{i, max}^2 - \lVert \boldsymbol{T_i} \rVert^2)- 
    \tfrac{\gamma_{T,i}}{(T_{i, max}^2 -  \lVert \boldsymbol{T_i} \rVert^2)}.
\end{split}
\end{equation}

\par
Based on the linear inequality (\ref{eqn: linear ineq for QP}), we formulate the QP expressing the saturation filter as follows:
\begin{equation}
\begin{aligned} \label{eqn: QP for saturation filter}
\ddot{\boldsymbol{\xi}}_2^* = \arg \min_{\ddot{\boldsymbol{\xi}}_2} &\tfrac{1}{2} \lVert \ddot{\boldsymbol{\xi}}_2 - \ddot{\boldsymbol{\xi}}_{nom} \rVert_{\boldsymbol{W}_s}^2 \\ 
\text{s.t.} \quad & (\ref{eqn: linear ineq for QP})
\end{aligned}
\end{equation}
where $\boldsymbol{W}_s \in \mathbb{R}^{6 \times 6}$ is a weighting matrix.
The jerk $\ddot{\boldsymbol{\xi}}_2^*$ resulting from (\ref{eqn: QP for saturation filter}) is used to update desired acceleration $\dot{\boldsymbol{\xi}}_d$, twist $\boldsymbol{\xi}_d$, and pose $\boldsymbol{T}_d$.

\subsection{Haptic Feedback Wrench Generation}
\par
Since the master device is driven at the acceleration/wrench level, the corrections implemented by the interaction and saturation filters must be converted to an acceleration/wrench signal.
Recall from (\ref{eqn: affine mapping 1}) that $\boldsymbol{G}$ maps a spatial acceleration to the corresponding control wrench expressed in the body frame $\mathcal{B}_1$. We therefore express each filter correction as an acceleration cue and render it through the same $\boldsymbol{G}$.
The velocity-level adjustment made by the interaction filter is transformed into feedback wrench $\boldsymbol{w}_{filt, 1} \in \mathbb{R}^6$ based on an acceleration cue $\boldsymbol{a}_1 \in \mathbb{R}^6$:
\begin{equation} \label{eqn: interaction filter feedback}
\boldsymbol{w}_{filt, 1} = \boldsymbol{G} \boldsymbol{a}_1, \qquad \boldsymbol{a}_1 \triangleq \tfrac{1}{T_h} (\boldsymbol{\xi}_1^* - \boldsymbol{\xi}_c)
\end{equation}
where $T_h > 0$ is a time constant.
Similarly, the jerk-level modification of the saturation filter is converted to feedback wrench $\boldsymbol{w}_{filt, 2} \in \mathbb{R}^6$ as follows:
\begin{equation} \label{eqn: saturation filter feedback}
\boldsymbol{w}_{filt, 2} = \boldsymbol{G} \boldsymbol{a}_2, \qquad \boldsymbol{a}_2 \triangleq T_h (\ddot{\boldsymbol{\xi}}_2^* - \ddot{\boldsymbol{\xi}}_{nom}).
\end{equation}

\par
The feedback wrenches $\boldsymbol{w}_{filt, 1}$ and $\boldsymbol{w}_{filt, 2}$ are linearly summed to form the total feedback wrench $\boldsymbol{w}_{filt}$:
\begin{equation} \label{eqn: total feedback wrench}
\boldsymbol{w}_{filt} = \boldsymbol{K}_1 \boldsymbol{w}_{filt, 1} + \boldsymbol{K}_2 \boldsymbol{w}_{filt, 2}.
\end{equation}
Here, $\boldsymbol{K}_1$ and $\boldsymbol{K}_2 \in \mathbb{R}^{6 \times 6}$ are user-defined positive diagonal matrices, with $\boldsymbol{K}_2>\boldsymbol{K}_1$ to emphasize the saturation feedback.

\section{Experimental Results}
\label{sec5}
\subsection{Experimental Setup}
The master device is a robotic manipulator based on the FACTR hardware~\cite{liu2025factr}. The slave device is a hexarotor-based omnidirectional tiltrotor, hereafter simply referred to as the AM. The AM is equipped with KDE2315XF-965 brushless motors, APC 6$\times$4$\times$3.2 propellers, and Dynamixel XM430-W210-T servo motors. An RGB camera, oCam-5CRO-U-M, is mounted to provide visual feedback, and a 0-DoF manipulator is rigidly attached to the vehicle for manipulation. The flight controller is a Pixhawk~6C running PX4, and an ASUS NUC13 serves as the onboard computer running Robot Operating System~2 (ROS~2) to control the actuators. State estimation is obtained by fusing motion-capture measurements from an OptiTrack system with IMU data from the Pixhawk. The overall control loop runs at $1/\Delta T=200$~Hz and the filter implementation uses finite-difference approximations with sampling period $\Delta T$; accordingly, we set $T_{c_1}=T_{c_2}=T_h=\Delta T$ and propagate the desired states via discrete-time numerical integration.

Unless otherwise specified, the position-error bound is set to $e_{p,i}^{[M]}=0.025$~m for $i\in\{x,y,z\}$ and the attitude-error bound is set to $\theta_R=5^\circ$. The omnidirectionality test uses a relaxed attitude bound of $\theta_R=10^\circ$. The feedback gains in (\ref{eqn: total feedback wrench}) are fixed as $\boldsymbol{K}_1=\mathrm{diag}([0.02,\,0.02,\,0.02,\,0.01,\,0.01,\,0.01])$ and $\boldsymbol{K}_2=5\boldsymbol{K}_1$. In Section~V.C, $\boldsymbol{K}_1$ and $\boldsymbol{K}_2$ are increased by a factor of 5 to improve the visibility of the rendered haptic feedback.

\subsection{Comparative Experiments on Haptic Feedback from AM}

\begin{figure}[!t]
    \centering
    \begin{subfigure}[b]{0.48\linewidth}
        \centering
        \includegraphics[width=\linewidth]{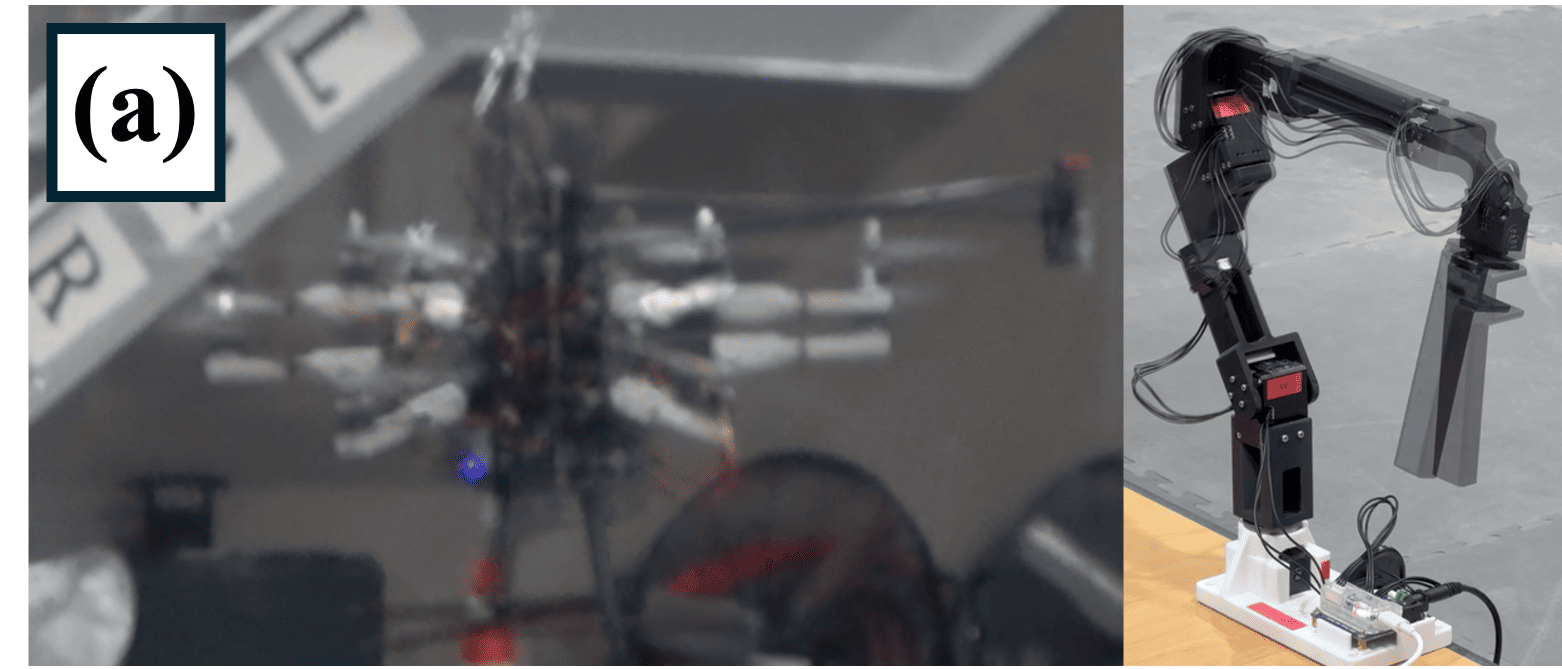}
        \label{fig:freeflight_dob}
    \end{subfigure}
    \hfill
    \begin{subfigure}[b]{0.48\linewidth}
        \centering
        \includegraphics[width=\linewidth]{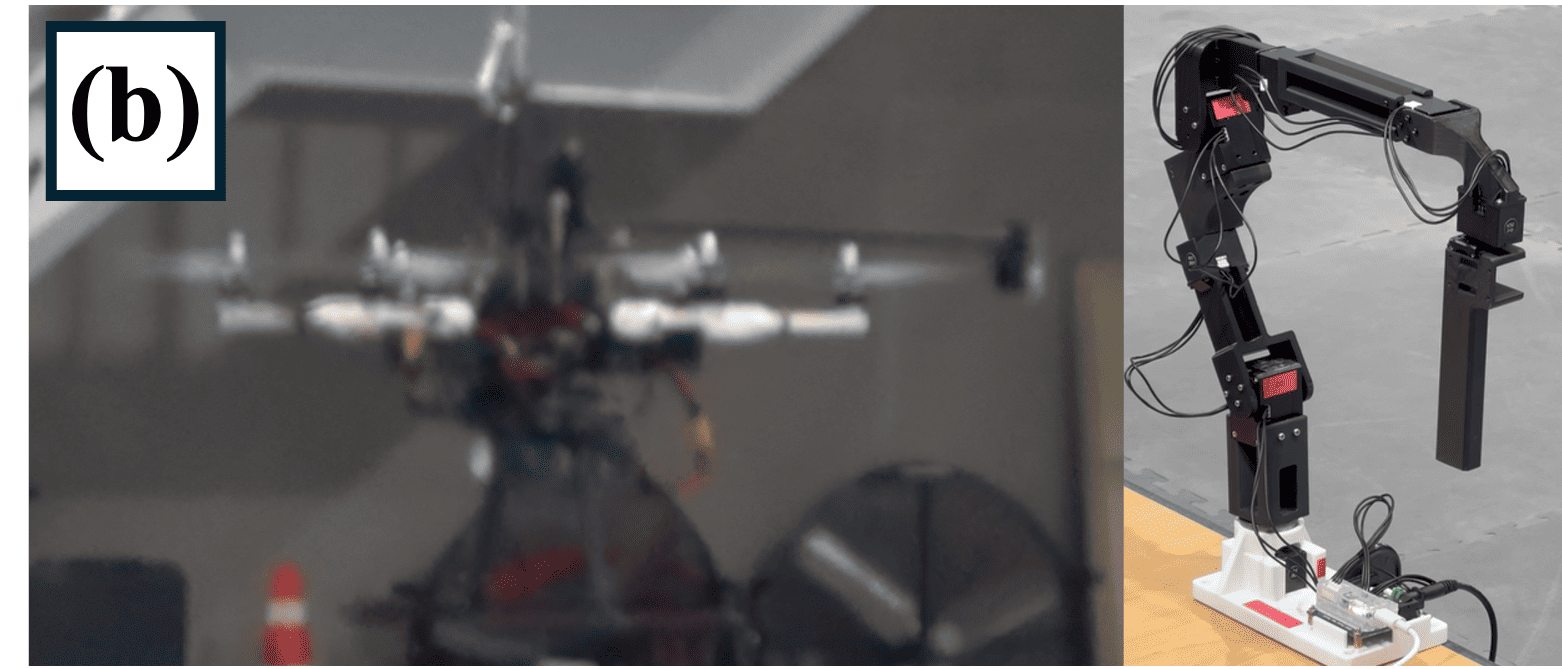}
        \label{fig:freeflight_filter}
    \end{subfigure}
    \vspace{-5mm}
    \caption{The behaviors of the AM and master device during free-flight using haptic feedback generated by (a) the momentum-based wrench estimation~\cite{lee2020aerial,kim2017robust} and (b) the proposed method. Note that (a) exhibits unintended drift while hovering, whereas (b) remains stable.}
    \label{fig:freeflight_compare}
\end{figure}

\begin{figure}[!t]
    \centering
    \vspace{-2mm}    \includegraphics[width=1.0\linewidth]{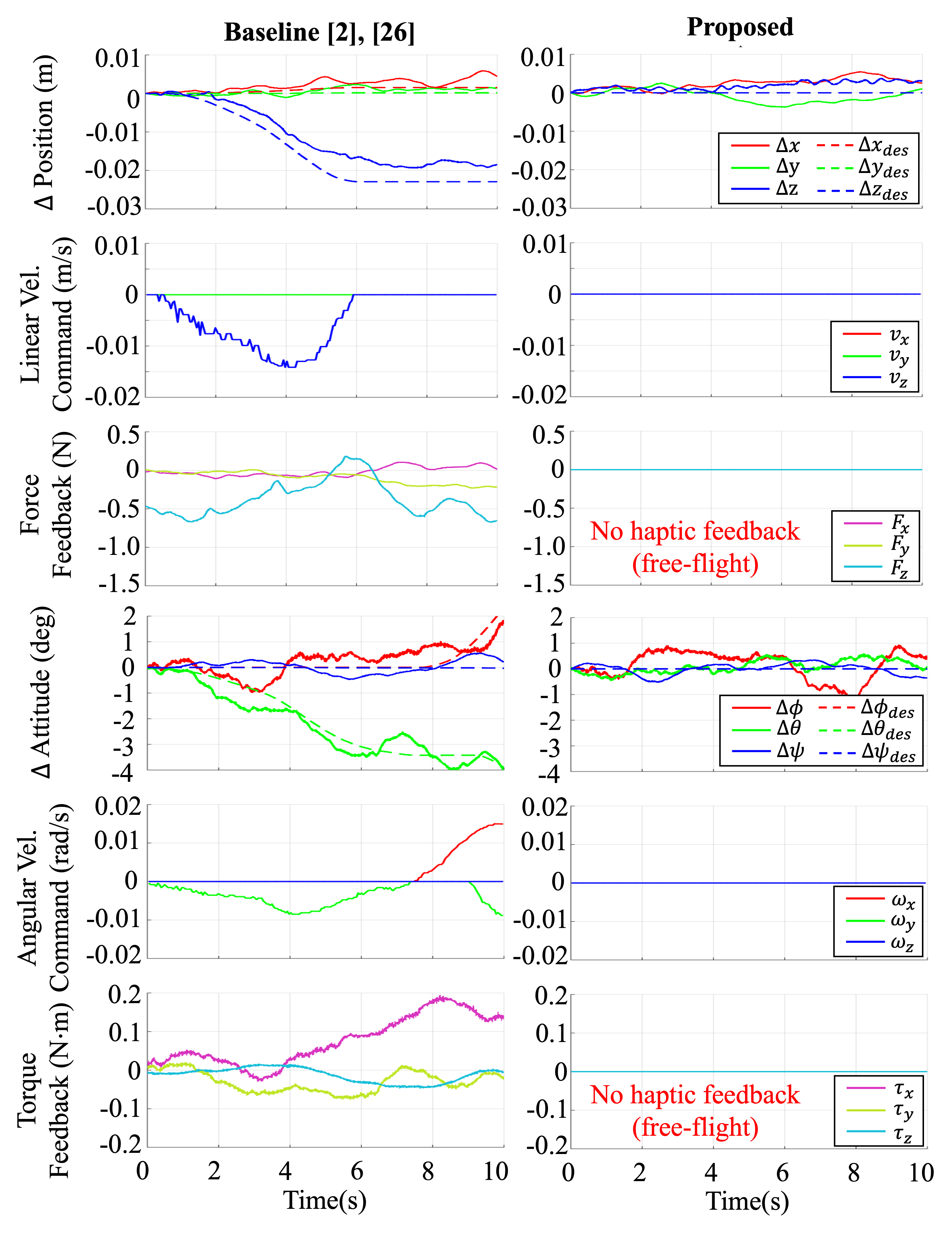}
    \vspace{-7mm}
    \caption{Comparison between the momentum-based wrench estimation and the proposed haptic feedback method. In this figure, we let $\Delta$ denote the deviation from the initial value, and $\phi$, $\theta$, and $\psi$ represent the roll, pitch, and yaw angles of the AM, respectively.}
    \label{fig:dob_filter_comparison}
\end{figure}

In this section, we compare our proposed FT-sensorless method for generating haptic feedback with a momentum-based wrench estimation baseline under nominal hovering conditions, as illustrated in Fig.~\ref{fig:freeflight_compare}. In our implementation of the baseline, we adopt different disturbance-observer (DOB) designs for translation and rotation, following~\cite{lee2020aerial} and~\cite{kim2017robust}, respectively. To isolate the feedback generated solely by the slave side, the experiments are conducted without the operator interacting with the master device. When the haptic feedback is calculated from the momentum-based wrench estimation, the average estimated wrench under the nominal hovering condition is first computed, then only the deviation from this average value is fed back to the master device. 

As shown in Fig.~\ref{fig:dob_filter_comparison}(a), with the baseline method, it is noted that the estimated wrench drifts over time even during the hovering condition. Accordingly, unintended linear and angular velocity commands at the master device are generated so that the AM keeps deviating from the hovering condition, thereby further amplifying the estimation drift. In situations when an operator is present, such drifting feedback would continuously disturb the operator even during the free-flight (i.e., even in the absence of physical interaction between the AM and its surroundings). This is because the momentum-based wrench estimation interprets the interaction wrench as a lumped disturbance including the wrench from physical interaction, aerodynamic effects, and model uncertainties.

\begin{figure}[!t]
    \centering
\includegraphics[width=0.64\linewidth]{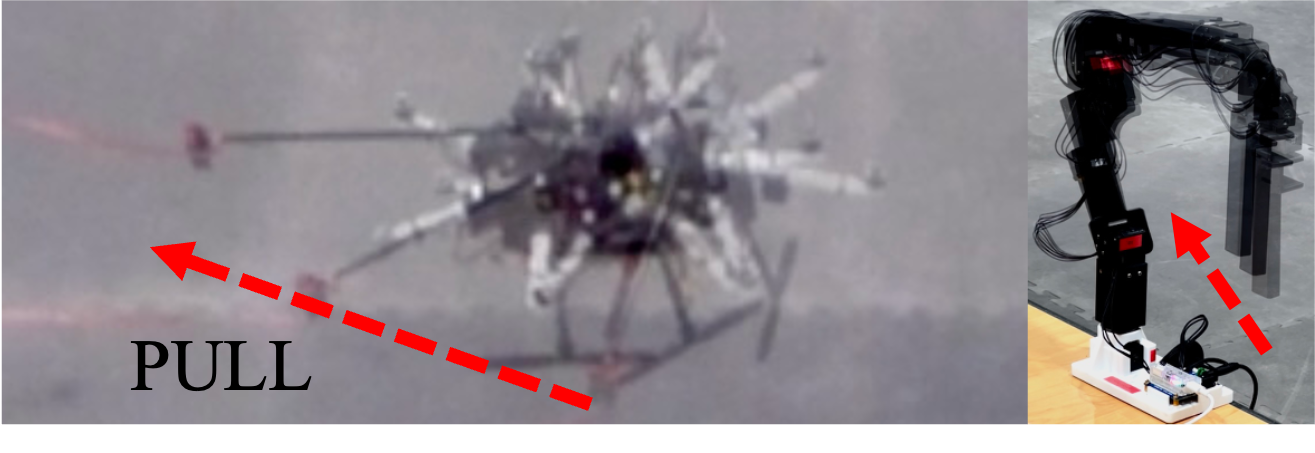}
\vspace{-2mm}
    \caption{The behaviors of the AM and master device when the AM is manually displaced from its desired pose by pulling. It can be observed that the end-effector of the master device moves in accordance with the AM's deviation from the desired pose.}
    \label{fig:feedback}
\end{figure}

\begin{figure}[!t]
    \centering
    \vspace{-2mm}
    \includegraphics[width=1.0\linewidth]{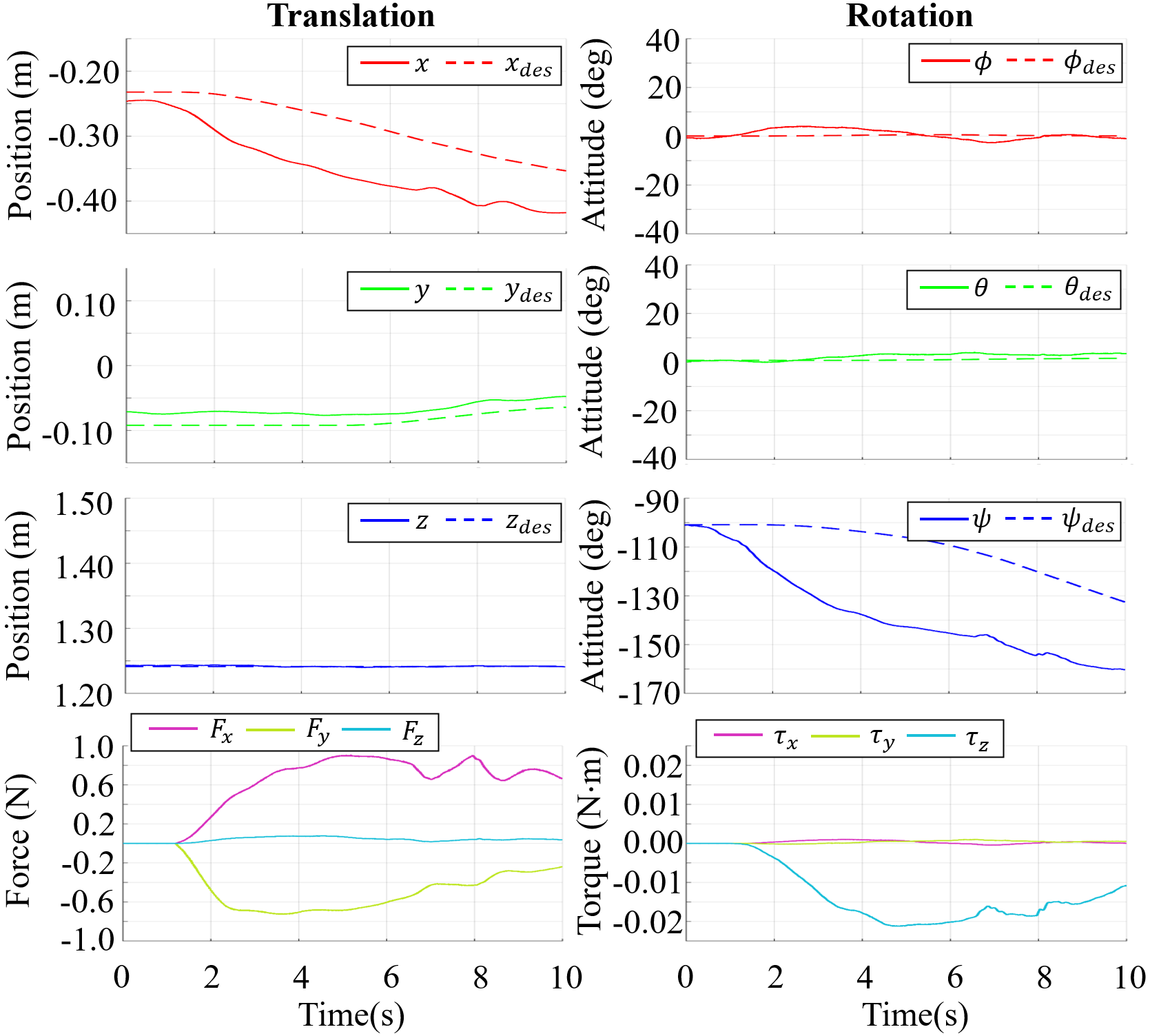}
    \vspace{-6mm}
    \caption{Measured wrench feedback under externally induced AM pose deviations. In this figure, $\phi$, $\theta$, and $\psi$ represent the roll, pitch, and yaw angles of the AM, respectively.}
    \label{fig:feedback_plot}
\end{figure}

By contrast, the proposed method generates wrench feedback only when the tracking error exceeds a prescribed threshold. Therefore, under nominal hovering conditions, where the tracking error remains within the allowable bound, no wrench feedback is transmitted to the operator. This prevents unintended feedback during free-flight while preserving haptic feedback during physical interaction.
Moreover, the proposed feedback method allows the operator to perceive external disturbances acting on the AM, such as sudden wind gusts or impacts, even when the system is not engaged in a manipulation task, extending the transparency of the teleoperation framework.

\vspace{-1mm}

\subsection{Validation of Proposed Feedback}

To validate the proposed feedback method, we examine the behavior of the master device when the slave is externally displaced from its desired pose as shown in Fig.~\ref{fig:feedback}. In Fig.~\ref{fig:feedback_plot}, we can observe that the proposed method generates wrench feedback that aligns with the direction of the induced deviation. This result demonstrates that the proposed approach effectively conveys external interaction cues to the master device in an intuitive and directionally coherent manner.

\begin{figure}[!t]
    \centering
    \includegraphics[width=1.0\linewidth]{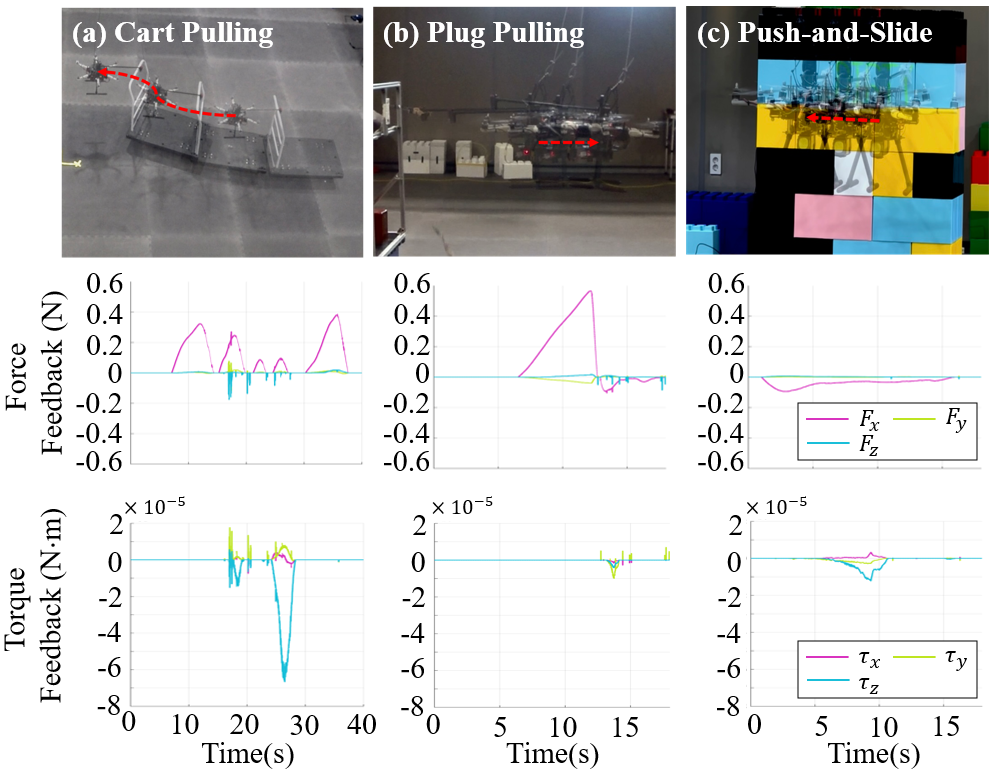}
    \vspace{-6mm}
    \caption{Snapshots and wrench feedback results from (a) cart pulling, (b) plug pulling, and (c) push-and-slide.}
    \label{fig:tasks}
\end{figure}

\begin{figure}[!t]
    \centering
    \vspace{-1mm}    \includegraphics[width=1.0\linewidth]{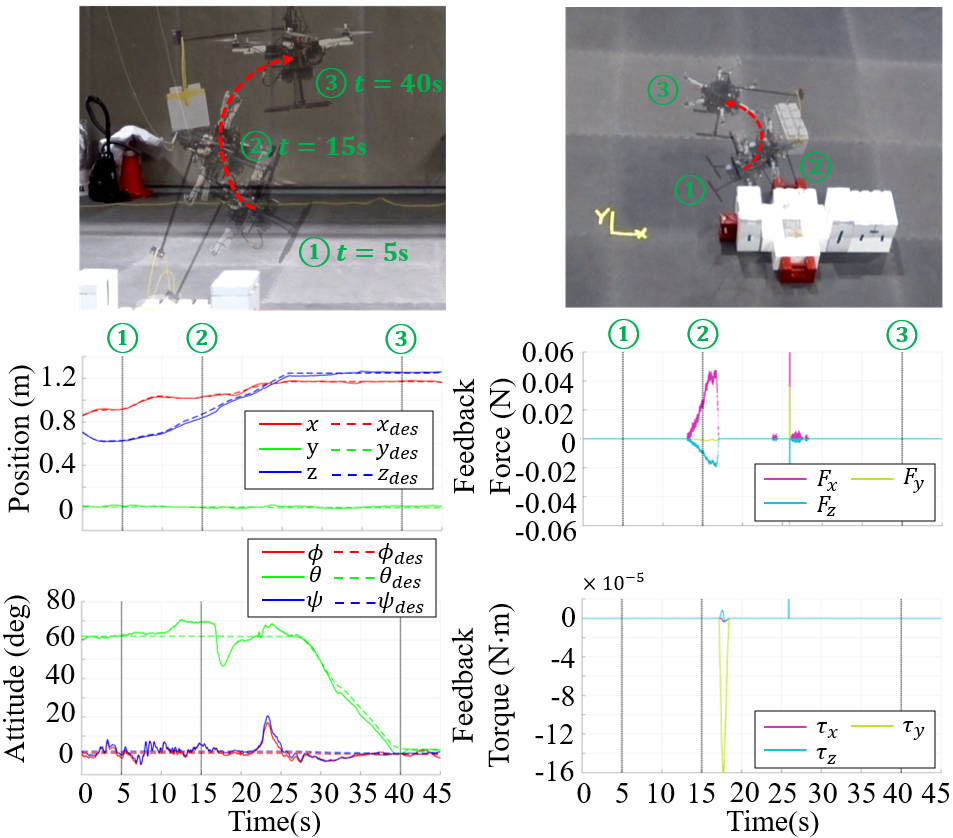}
    \vspace{-5mm}
    \caption{The AM lifting up the box during the pick-and-place task, along with the corresponding pose and haptic feedback wrenches.}
    \label{fig:pick_and_place}
\end{figure}

\vspace{-1mm}

\subsection{Demonstration of Various Tasks}
To evaluate our teleoperation framework in diverse scenarios involving static and dynamic physical interactions, we conduct three tasks---cart pulling, plug pulling, and push-and-slide---as shown in Fig.~\ref{fig:tasks}.
We observe that the feedback wrench is aligned with the expected directions of physical interaction during manipulation. The successful completion of these tasks demonstrates that the proposed framework can reliably convey interaction cues to the operator across different manipulation scenarios.

\subsection{Validation of Omnidirectionality}
To validate the omnidirectionality of the AM, we conduct a pick-and-place task. The operator must approach and pick the target object in a near-vertical direction since the object (handled box) is placed between two tall obstacles, as shown in Fig.~\ref{fig:pick_and_place}. The AM successfully hooks and lifts up the box, and then transports it to the takeoff position.

\subsection{Ablation Study on Saturation Filter}
\begin{figure}[!t]
    \centering
    \includegraphics[width=1.0\linewidth]{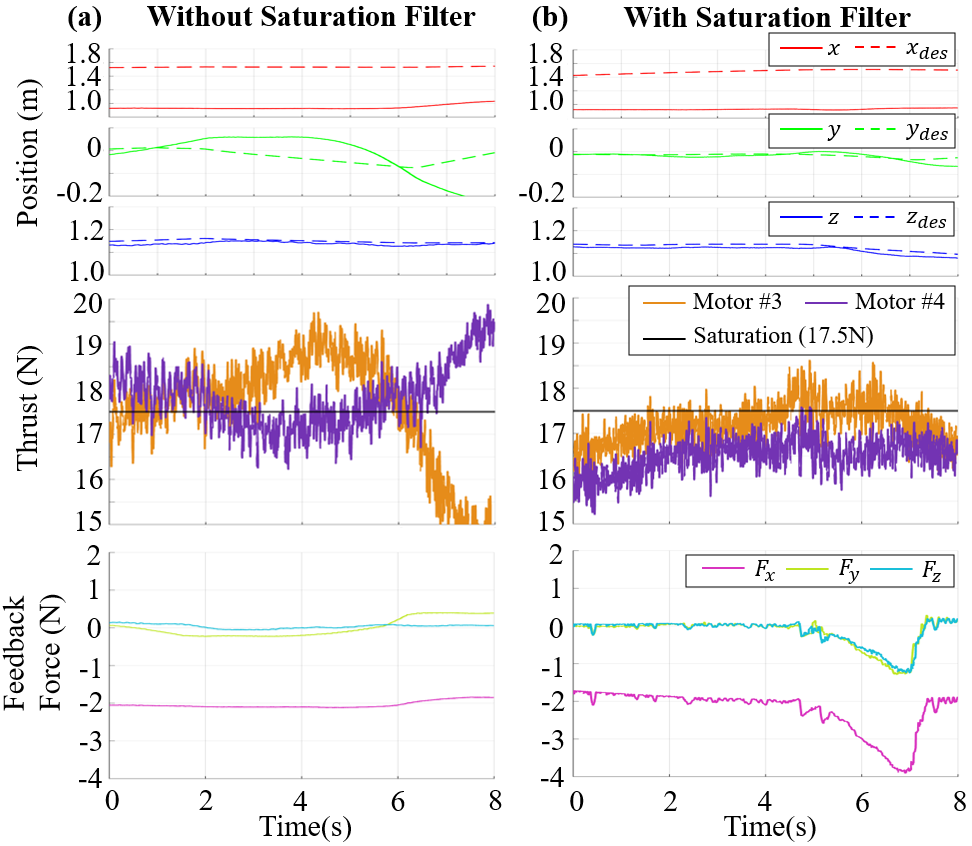}
    \vspace{-6mm}
    \caption{Ablation study on the saturation filter conducted with a static pushing scenario.}
    \label{fig:ablation}
\end{figure}
To show the efficacy of considering motor saturation, we conduct an ablation study on the saturation filter with a static pushing scenario, which may readily violate thrust limits. For both experiments with and without the saturation filter, the interaction filter is enabled and the operator commanded a constant linear velocity of 0.15~m/s in the $x$ direction. As shown in Fig.~\ref{fig:ablation}(a), when the saturation filter is disabled, the sustained velocity command along the $x$ direction drives motors \#3 and \#4 into saturation. As a result, a substantial deviation in the $y$ direction occurs, which in turn destabilizes the AM. In contrast, in Fig.~\ref{fig:ablation}(b), we can observe that the saturation filter successfully constrains the motor thrusts near the saturation region. It is mentioned that the saturation threshold is set with a safety margin, because strict enforcement of the hard bound may not always be guaranteed due to discretization in implementation.

From the perspective of haptic feedback, Fig. \ref{fig:ablation}(b) also indicates that the saturation filter generates force feedback in the $-x$, $-y$, and $-z$ directions, implying that the desired position of the AM was adjusted in those directions. Also, from the thrust of motor \#3, which operates closest to saturation since it is located at the front-right side of the AM, we can observe that our saturation filter seamlessly shifts the desired position slightly toward the $-x$, $-y$, and $-z$ directions to alleviate the load on this motor while preserving flight stability. These results demonstrate that the saturation filter prevents motor saturation and destabilization while providing informative cues on the direction of the corrective action, guiding the operator away from unsafe commands.

\section{Conclusions}
\label{sec6}
In this work, we present a force-sensorless bilateral teleoperation framework for omnidirectional aerial manipulators that integrates visual and haptic feedback. A two-stage hierarchical safety filter enforces tracking error and motor thrust constraints and maps filter induced command adjustments into haptic cues, enabling perception of both physical interaction and impending motor saturation without onboard FT sensors. Hardware experiments show that, compared to momentum-based wrench estimation, the proposed approach suppresses unintended feedback during  free flight while remaining responsive and directionally consistent under interaction, enabling multiple manipulation tasks. The ablation study further confirms that the saturation filter mitigates motor saturation and improves stability while providing informative corrective feedback. Future work will focus on formal stability analysis and systematic user studies.

\section*{Acknowledgment} The authors would like to thank Yeonjoon Kim for designing the hexarotor-based omnidirectional tiltrotor platform.

\bibliographystyle{IEEEtran}
\bibliography{root}

\end{document}